\pdfoutput=1
\documentclass[11pt]{article}
\usepackage{acl2023}

\usepackage[utf8]{inputenc}

\usepackage{inconsolata}

\usepackage{times}
\usepackage{latexsym}
\usepackage{xspace}
\usepackage{enumitem}

\usepackage{url}
\usepackage{xcolor}
\usepackage{booktabs}
\usepackage{diagbox}
\usepackage{multirow}
\usepackage{balance}
\usepackage{amsfonts}
\usepackage{amsmath}
\usepackage{amssymb}
\usepackage{graphicx}
\usepackage{subfigure}
\usepackage{microtype}
\usepackage{wrapfig,lipsum}
\usepackage{algorithm}
\usepackage{algorithmicx}
\usepackage{algpseudocode}
\usepackage{makecell}
\usepackage[T1]{fontenc}
\usepackage{soul}
\usepackage{tabularx, booktabs}
\usepackage{todonotes}
\usepackage{comment}
\newcommand{\cbb}{\mathbf{c}}
\newcolumntype{Y}{>{\centering\arraybackslash}X}
\newcolumntype{L}{>{\arraybackslash}X}
\newcolumntype{M}[1]{>{\centering\arraybackslash}m{#1}}

\algnewcommand\algorithmicinput{\textbf{Input:}}
\algnewcommand\algorithmicoutput{\textbf{Output:}}
\algnewcommand\algorithmicparameter{\textbf{Parameters:}}
\algnewcommand\INPUT{\item[\algorithmicinput]}
\algnewcommand\OUTPUT{\item[\algorithmicoutput]}
\algnewcommand\PARAMETER{\item[\algorithmicparameter]}

\newcommand{\clstoken}{\texttt{[CLS]}}

\newcommand{\ourmodel}{\texttt{TASER}}

\newcommand{\eat}[1]{\ignorespaces}
\input{Definitions.tex}

\usepackage{todonotes}
\makeatletter
\newcommand*\iftodonotes{\if@todonotes@disabled\expandafter\@secondoftwo\else\expandafter\@firstoftwo\fi}  %
\makeatother

\newcommand\blfootnote[1]{%
  \begingroup
  \renewcommand\t
  \addtocounter{footnote}{-1}%
  \endgroup
}

\title{
Task-Aware Specialization for Efficient and Robust Dense Retrieval for Open-Domain Question Answering
}

\author{
Hao Cheng\textsuperscript{$\spadesuit$}
\quad
Hao Fang\textsuperscript{$\clubsuit$}
\quad
Xiaodong Liu\textsuperscript{$\spadesuit$}
\quad
Jianfeng Gao\textsuperscript{$\spadesuit$}
\\ 
\textsuperscript{$\spadesuit$} Microsoft Research 
\quad
\textsuperscript{$\clubsuit$} Microsoft Semantic Machines
\\
{\tt \{chehao,hafang,xiaodl,jfgao\}@microsoft.com}
}

\date{}

\begin{document}
\maketitle

\begin{abstract}
Given its effectiveness on knowledge-intensive natural language processing tasks, dense retrieval models have become increasingly popular.
Specifically, the \textit{de-facto} architecture for open-domain question answering uses two isomorphic encoders that are initialized from the same pretrained model but separately parameterized for questions and passages.
This bi-encoder architecture is parameter-inefficient in that there is no parameter sharing between encoders.
Further, recent studies show that such dense retrievers underperform BM25 in various settings.
We thus propose a new architecture, {\bf T}ask-{\bf A}ware {\bf S}pecialization for d{\bf E}nse {\bf R}etrieval (\ourmodel), which enables parameter sharing by interleaving shared and specialized blocks in a single encoder.
Our experiments on five question answering datasets show that \ourmodel\ can achieve superior accuracy, surpassing BM25, while using about $60\%$ of the parameters as bi-encoder dense retrievers.
In out-of-domain evaluations, \ourmodel\ is also empirically more robust than bi-encoder dense retrievers.
Our code is available at \url{https://github.com/microsoft/taser}.
\end{abstract}

\section{Introduction}
\label{sec:intro}
Empowered by learnable neural representations built upon pretrained language models, the dense retrieval framework has become increasingly popular for fetching external knowledge in various natural language processing tasks \cite{lee-etal-2019-latent,guu2020realm,lewis2020retrievalaugmented}.
For open-domain question answering (ODQA), 
the \textit{de-facto} dense retriever is the bi-encoder architecture \cite{lee-etal-2019-latent,karpukhin-etal-2020-dense},
consisting of a question encoder and a passage encoder.
Typically, the two encoders are isomorphic but separately parameterized, 
as they are initialized from the same pretrained model and then fine-tuned 
on the task.

Despite of its popularity, this bi-encoder architecture with fully decoupled parameterization has some open issues.
First, from the efficiency perspective, the bi-encoder parameterization apparently results in scaling bottleneck for both training and inference.
Second, empirical results from recent studies show that such bi-encoder dense retrievers underperform its sparse counterpart BM25 \cite{bm25} in various settings.
For example, both \citet{lee-etal-2019-latent} and \citet{karpukhin-etal-2020-dense} suggest the inferior performance on SQuAD \cite{squad1} is partially due to the high lexical overlap between questions and passages, which gives BM25 a clear advantage.
\citet{sciavolino-etal-2021-entityq} also find that bi-encoder dense retrievers are more sensitive to distribution shift 
than BM25,
resulting in poor generalization on questions with rare entities.

In this paper, we develop {\bf T}ask-{\bf A}ware {\bf S}pecial-ization for d{\bf E}nse {\bf R}etrieval, \ourmodel,
as a more parameter-efficient and robust architecture.
Instead of using two isomorphic and fully decoupled Transformer \cite{vaswani2017attention} encoders, \ourmodel\ interleaves shared encoder blocks with specialized ones in a single encoder,
motivated by recent success in using Mixture-of-Experts (MoE) to scale up Transformer \cite{fedus-et-al-2021-switchtransformer}.
For the shared encoder block, the entire network is used to encode both questions and passages.
For the specialized encoder block, some sub-networks are task-specific and activated only for certain encoding tasks.
To choose among task-specific sub-networks, \ourmodel\ uses an input-dependent \textit{routing mechanism}, \ie questions and passages are passed through separate dedicated sub-networks.

We carry out both in-domain and out-of-domain evaluation for \ourmodel.
For the in-domain evaluation, we use five popular ODQA datasets.
Our best model outperforms BM25 and existing bi-encoder dense retrievers, while using much less parameters.
It is worth noting that \ourmodel\ can effectively close the performance gap on SQuAD between dense retrievers and BM25.
One interesting finding from our experiments is that excluding SQuAD from the multi-set training is unnecessary,
which was a suggestion made by \citet{karpukhin-etal-2020-dense} and adopted by most follow-up work.
Our out-of-domain evaluation experiments use EntityQuestions \cite{sciavolino-etal-2021-entityq} and BEIR \cite{thakur-et-al-2021-beir}.
Consistent improvements over the doubly parameterized bi-encoder dense retriever are observed in these zero-shot evaluations as well.
Our code is available at \url{https://github.com/microsoft/taser}.

\section{Background}
\label{sec:biencoder}

In this section, we provide necessary background about the bi-encoder architecture for
dense passage retrieval which is widely used in ODQA \cite{lee-etal-2019-latent,karpukhin-etal-2020-dense}
and is the primary baseline model in our experiments.

The bi-encoder architecture consists of a question encoder and a passage encoder, 
both of which are usually Transformer encoders \cite{vaswani2017attention}.
A Transformer encoder is built up with a stack of Transformer blocks.
Each block consists of a multi-head self-attention (MHA) sub-layer and a feed-forward network (FFN) sub-layer,
with residual connections \cite{he-etal-2016-resnet} and layer-normalization \cite{ba-etal-2016-layer-norm} 
applied to both sub-layers.
Given an input vector $\hvec \in \mathbb{R}^d$, the FFN sub-layer produces an output vector as following
\begin{align}
    \texttt{FFN}(\hvec) = \textbf{W}_2 \max\{ 0, \textbf{W}_1 \hvec + \bvec_1 \} + \bvec_2,
    \label{eq:ffn}
\end{align}
where $\textbf{W}_1\in\RR^{m \times d}, \textbf{W}_2\in\RR^{d \times m}, \bvec_1\in\RR^m$, 
and $\bvec_2\in\RR^d$ are learnable parameters.
For a sequence of $N$ tokens, each Transformer block produces $N$ corresponding vectors, 
together with a vector for the special prefix token \clstoken\ which can be used as the representation 
of the sequence.
We refer readers to \cite{vaswani2017attention} for other details about Transformer.
Typically the question encoder and passage encoder are initialized from a pretrained language model 
such as BERT \cite{devlin-etal-2019-bert}, but they are parameterized separately, 
\ie their parameters would differ after training.

The bi-encoder model independently encodes questions and passages into $d$-dimension vectors, 
using the final output vectors for \clstoken\ from the corresponding encoders,
denoted as $\qvec \in \mathbb{R}^d$ and $\pvec \in \mathbb{R}^d$, respectively.
The relevance between a question and a passage can then be measured in the vector space using dot product, \ie
$\text{sim}(\qvec, \pvec) = \qvec^T\pvec$.
During training, the model is optimized based on a contrastive learning objective,
\begin{align}
    L_{sim} = -{\exp(\text{sim}(\qvec, \pvec^+)) \over \sum_{\pvec^\prime\in\mathcal{P}\cup\{\pvec^+\}}  \exp(\text{sim}(\qvec, \pvec^\prime))},
    \label{eqn:obj_sim}
\end{align}
where $\pvec^+$ is the relevant (positive) passage for the given question, 
and $\mathcal{P}$ is the set of irrelevant (negative) passages.
During inference, all passages are pre-converted into vectors using the passage encoder.
Then, each incoming question is encoded using the question encoder, and a top-$K$ list of most relevant passages are retrieved based on their relevance scores with respect to the question.

Although the bi-encoder dense retrieval architecture has achieved impressive results in ODQA, few work has attempted to improve its parameter efficiency.
Further, compared to the spare vector space model BM25 \cite{bm25}, such bi-encoder dense retrievers sometimes suffer from inferior generalization performance, \eg when the training data is extremely biased \cite{lebret-etal-2016-neural,karpukhin-etal-2020-dense} 
or when there is a distribution shift \cite{sciavolino-etal-2021-entityq}.
In this paper, we conjecture that the unstable generalization performance is partially related to the unnecessary number of learnable parameters in the model.
Therefore, we develop a task-aware specialization architecture for dense retrieval with parameter sharing 
between the question and passage encoders, which turns out to improve both parameter efficiency and generalization performance.
\section{Proposed Model: \ourmodel}
\label{sec:taser}

\begin{figure}[t]
    \centering
    \includegraphics[width=0.4\textwidth]{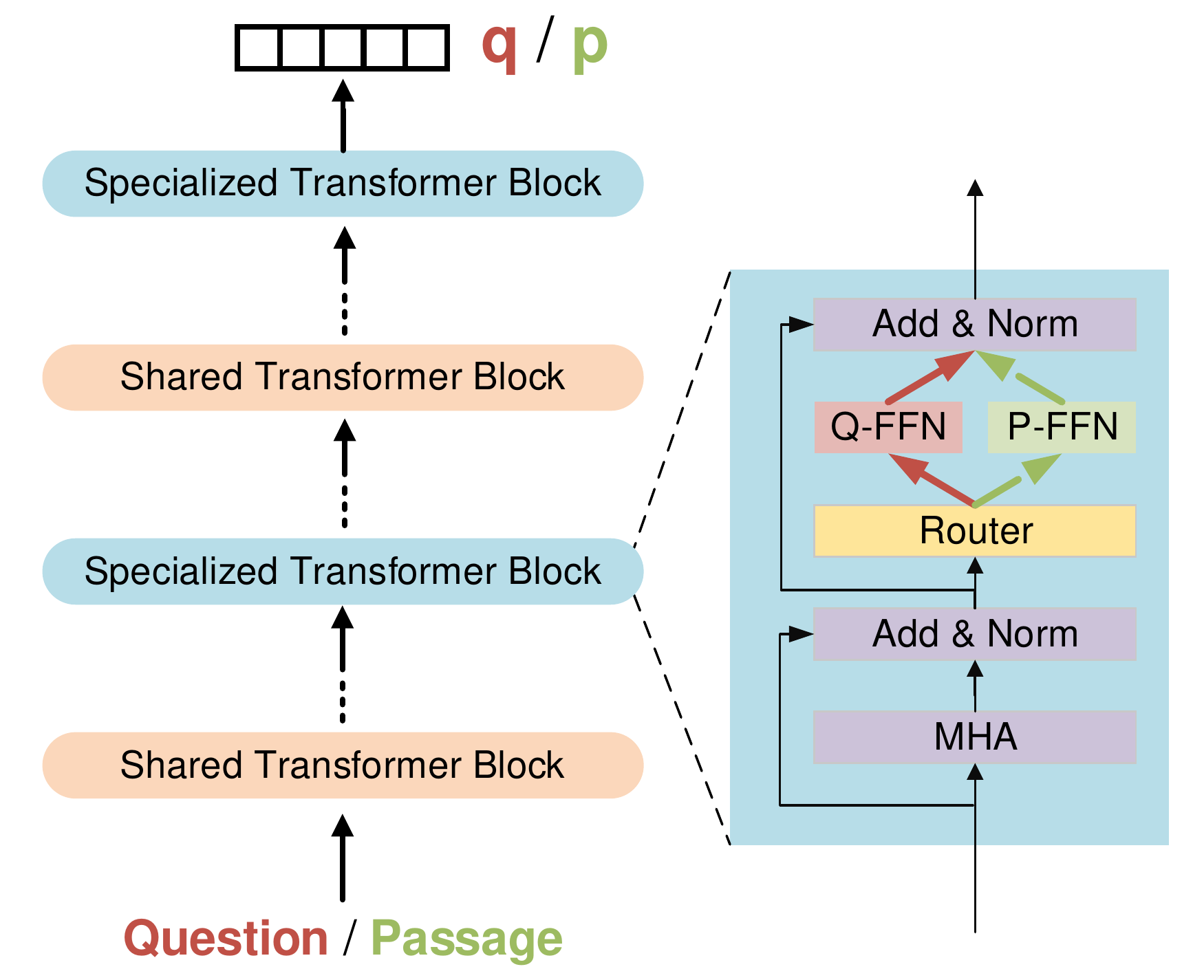}
    \caption{The architecture overview of \ourmodel.
    The specialized transformer block consists of a Q-FFN for questions, a P-FFN for passages, and a router which deterministically chooses between these two expert FFN sub-layers based on input.}
    \label{fig:taser}
\end{figure}

As shown in \autoref{fig:taser},
\ourmodel\ interleaves shared Transformer blocks 
with specialized ones.
The shared Transformer block is identical to the Transformer block used in the bi-encoder architecture, 
but the entire block is shared for both questions and passages.
In the specialized block, we apply MoE-style task-aware specialization to the FFN sub-layer, following \cite{fedus-et-al-2021-switchtransformer},
where the router always routes the input to a single expert FFN sub-layer.
In our experiments, we use a simple yet effective routing mechanism,
which uses an expert sub-layer (Q-FFN) for questions and another (P-FFN) for passages.
The router determines the expert FFN sub-layer based on whether the input is a question or a passage.
Other routing mechanisms are discussed in Appendix~\ref{appendix:routing_mechanisms}.

\ourmodel\ uses one specialized Transformer block after every $T$ shared Transformer blocks in the stack, 
starting with a
shared one at the bottom.
Our preliminary study indicates that the model performance is not sensitive to the choice of $T$, 
so we use $T = 2$ for experiments in this paper.

Similar to the bi-encoder architecture, 
\ourmodel\ is trained using the contrastive learning objective $L_{sim}$ defined in \autoref{eqn:obj_sim}.
Specifically, the objective needs to use a set of negative passages $\mathcal{P}$ for each question.
Following \citet{xiong-etal-2020-iter-train} and \citet{qu-etal-2020-rocketqa},
we construct $\mathcal{P}$ via hard negatives mining (Appendix~\ref{sec:hard_negatives_mining}).
Our experiments use the \textit{multi-set} training paradigm,
\ie the model is trained by combining data from multiple datasets
to obtain a model that works well across the board.

\section{Experiments}
\label{sec:experiment}

\subsection{In-Domain Evaluation}
\label{ssec:multi_set}

\begin{table}[t!]
\small
\centering

\begin{tabular}{l|c|c|c|c|c}
\toprule

& \textbf{NQ}
& \textbf{TQ}
& \textbf{WQ}
& \textbf{CT} 
& \textbf{SQuAD}
\\
\midrule

BM25$^{(1)}$        
& 62.9 
& 76.4 
& 62.4 
& 80.7 
& 71.1 \\
\midrule

\multicolumn{6}{c}{\textbf{Multi-Set Training (without SQuAD)}} \\
\midrule

DPR$^{(1)}$   
& 79.5 
& 78.9 
& 75.0 
& 88.8 
& 52.0 \\

DPR\textsubscript{BM25}$^{(1)}$  
& 82.6 
& 82.6 
& 77.3 
& 90.1 
& 75.1 \\

xMoCo$^{(2)}$ 
& 82.5 
& 80.1 
& 78.2 
& 89.4 
& 55.9 \\

SPAR\textsubscript{Wiki}$^{(3)}$
& 83.0 
& 82.6 
& 76.0 
& 89.9 
& 73.0 \\

SPAR\textsubscript{PAQ}$^{(4)}$    
& 82.7 
& 82.5 
& 76.3 
& 90.3 
& 72.9 \\

\midrule

\multicolumn{6}{c}{\textbf{Multi-Set Training (with SQuAD)}} \\
\midrule

DPR$^\dag$
& 80.9 
& 79.6 
& 74.0 
& 88.0 
& 63.1 \\

DPR$^\diamond$
& 82.5
& 81.8
& 77.8
& 91.2
& 67.0 \\

DPR$^\star$
& 83.7
& 82.6
& 78.9
& 91.6
& 68.0 \\

\ourmodel$^\diamond$
& 83.6 %
& 82.0 %
& 77.9 %
& 91.1 %
& 69.7 %
\\

\ourmodel$^\star$
& 84.9
& 83.4 
& 78.9 
& 90.8
& 72.9 \\

\ourmodel$^\star_{\rm BM25}$
& {\bf 85.0} 
& {\bf 84.0} 
& {\bf 79.6} 
& {\bf 92.1} 
& {\bf 78.0} \\
\bottomrule

\end{tabular}

\caption{In-domain evaluation results (test set R@20).
$^{(1)}$: \cite{ma-etal-2021-dpr-replication}.
$^{(2)}$: \cite{karpukhin-etal-2020-dense}.
$^{(3)}$: \cite{yang-etal-2021-xmoco}.
$^{(4)}$: \cite{chen-etal-2022-spar}.
$_{\rm BM25}$: combined with BM25 scores.
$^\dag$: initialized from BERT-base and without hard negatives mining.
$^\diamond$: initialized from BERT-base. 
$^\star$: initialized from coCondenser-Wiki.
The last five models are trained with the same hard negatives mining.
}
\label{tab:in_domain_eval_short}

\end{table}

We carry out in-domain evaluations on five ODQA datasets:
NaturalQuestions \citep[NQ;][]{nq}, 
TriviaQA \citep[TQ;][]{joshi-etal-2017-triviaqa},
WebQuestions \citep[WQ;][]{berant-etal-2013-semantic},
CuratedTrec \citep[CT;][]{curatedtrec},
and SQuAD \cite{squad1}.
All data splits and the Wikipedia collection for retrieval used in our experiments are the same as \citet{karpukhin-etal-2020-dense}.
The top-$K$ retrieval accuracy (R@K) is used for evaluation,
which evaluates whether any gold answer string is contained in the top $K$ retrieved passages.

Besides BERT-base, coCondenser-Wiki \cite{gao-callan-2022-unsupervised} is also used to initialize \ourmodel\ models.
We further present results of hybrid models that linearly combine the dense retrieval scores with the BM25 scores.
See Appendix~\ref{appendix:in_domain_evaluatios_details} for details.
Evaluation results are summarized in \autoref{tab:in_domain_eval_short}.\footnote{%
We also report R@100 scores in \autoref{tab:in_domain_eval}
and corresponding model sizes in \autoref{tab:model_size}.}
Note that the last five models \autoref{tab:in_domain_eval_short} are trained with the same hard negatives mining.

All prior work excludes SQuAD from the multi-set training, as suggested by \citet{karpukhin-etal-2020-dense}.
We instead train models using all five datasets.
Specifically, 
we observe that this would not hurt the overall performance,
and it actually significantly improves the performance on SQuAD,
comparing DPR$^{(1)}$ with DPR$^\dag$.

Comparing models initialized from BERT-base,
\ourmodel$^\diamond$ significantly outperforms 
xMoCo \cite{yang-etal-2018-hotpotqa}
and is slightly better than DPR$^\diamond$, 
using around 60\% parameters. 
SPAR \cite{chen-etal-2022-spar} is also initialized from BERT-base, but it augments DPR with another dense lexical model trained on 
either Wikipedia or PAQ \cite{lewis-etal-2021-paq}, which doubles the model sizes (\autoref{tab:model_size}).
\ourmodel$^\diamond$ is mostly on par with SPAR-Wiki and SPAR-PAQ, except on SQuAD, but its model size is about a quarter of SPAR.

\citet{gao-callan-2022-unsupervised} has shown the coCodenser model outperforms DPR models initialized from BERT-base in the single-set training setting.
Here, we show that using coCondenser-Wiki for initialization is also beneficial for \ourmodel\ under the multi-set setting,
especially for SQuAD where R@20 is improved by 3.2 points.
Notably, SQuAD is the only dataset among the five where DPR underperforms BM25, 
due to its higher lexical overlap between questions and passages. 
Nevertheless, \ourmodel$^\star$ surpasses BM25 on all five datasets, and they are either on-par or better than 
state-of-the-art dense-only retriever models, demonstrating its superior parameter efficiency.

Consistent with previous work, combining BM25 with dense models can further boost the performance,
particularly on SQuAD.
However, the improvement is more pronounced on DPR as compared to \ourmodel$^\star$,
indicating that \ourmodel$^\star$ is able to capture more lexical overlap features.
Finally, \ourmodel$^\star_{\rm BM25}$ sets new state-of-the-art performance on all five ODQA datasets.

We also compare the computation time needed for one epoch of training and validation.
The baseline DRP model takes approximately 15 minutes, 
whereas \ourmodel{} takes 11 minutes (26\% improvement), both measured using 16 V100-32G GPUs.

\subsection{Out-of-Domain Evaluation}
\label{ssec:ood_eval}

\begin{table}[t!]
\footnotesize
\centering

\begin{tabular}{l|c|cccc}
\toprule
& R@20 & \multicolumn{4}{c}{nDCG@10} \\
& EQ & AA & DBP & FEV & HQA \\
\midrule

BM25                           
& 71.2
& 31.5 & 31.3 & 75.3 & 60.3 \\
\midrule

DPR\textsubscript{Multi}      
& 56.7
& 17.5 & 26.3 & 56.2 & 39.1 \\

\ourmodel$^\diamond$            
& 64.7
& 32.8 & 31.4 & 59.6 & 50.7 \\
\ourmodel$^\star$               
& 66.7
& 30.5 & 31.6 & 58.8 & 54.5 \\ 

\bottomrule
\end{tabular}

\caption{Out-of-domain evaluation results on EntityQuestions (R@20) and four BEIR datasets (nDCG@10). 
BM25 and DPR\textsubscript{Multi} results are from 
\cite{sciavolino-etal-2021-entityq} and \cite{thakur-et-al-2021-beir}.}
\label{tab:ood_eval_short}

\end{table}

We use two benchmarks to evaluate the out-of-domain generalization ability of 
\ourmodel$^\diamond$ and \ourmodel$^\star$ from \autoref{tab:in_domain_eval_short} .
EntityQuestions \citep[EQ;][]{sciavolino-etal-2021-entityq} is used to measure the model sensitivity to entity distributions,
as DPR is found to perform poorly on entity-centric questions containing rare entities.
BEIR \cite{thakur-et-al-2021-beir} is used to study the model generalization ability in other genres of information retrieval tasks.
Specifically, we focus on four datasets from BEIR where DPR underperforms BM25, \ie
ArguAna \citep[AA;][]{wachsmuth-etal-2018-retrieval}, DBPedia \citep[DBP;][]{hasibi-etal-2017-dbpedia}, FEVER \citep[FEV;][]{thorne-etal-2018-fever}, and HotpotQA \citep[HQA;][]{yang-etal-2018-hotpotqa}.
Results are summarized in \autoref{tab:ood_eval_short}.
For EntityQuestions, we report R@20 scores following \citet{sciavolino-etal-2021-entityq}.\footnote{The R@20 scores are averaged over all relations. More evaluation metrics are reported in \autoref{tab:ood_entityq}.} 
For BEIR datasets, nDCG@10 scores are used following \citet{thakur-et-al-2021-beir}.

On EntityQuestions, both \ourmodel$^\diamond$ and \ourmodel$^\star$ outperform the doubly parameterized DPR\textsubscript{Multi} \cite{karpukhin-etal-2020-dense},
with \ourmodel$^\star$ being slightly better.
Similar to the in-domain evaluation results, \ourmodel\ can effectively reduce the performance gap between the dense retrievers and BM25.
These results further support our hypothesis that more parameter sharing can improve the model robustness for dense retrievers.

On BEIR datasets, we also observe that \ourmodel\ models consistently improve over DPR\textsubscript{Multi} across the board.
Notably, \ourmodel$^\diamond$ and \ourmodel$^\star$ can actually match the performance of 
BM25 on ArguAna and DBpedia.
Interestingly, coCondenser pre-training has mixed results here, \ie
\ourmodel$^\star$ is only better than \ourmodel$^\diamond$ 
on HotpotQA and on par or worse on other datasets.

\section{Related Work}
\label{sec:related_work}

Recent seminal work on dense retrieval demonstrates its effectiveness using Transformer-based
bi-encoder models by 
either continual pre-training with an inverse cloze task \cite{lee-etal-2019-latent} 
or careful fine-tuning \cite{karpukhin-etal-2020-dense}.
One line of follow-up work improves dense retrieval models via various 
continual pre-training approaches \cite{guu2020realm,Chang2020Pre-training,izacard-etal-2021-contriever,gao-callan-2022-unsupervised,oguz-etal-2021-domain}.
Better contrastive learning objectives are also introduced
\cite{xiong-etal-2020-iter-train,qu-etal-2020-rocketqa,yang-etal-2021-xmoco}.
Motivated by the success of augmenting dense models with sparse models,
\citet{chen-etal-2022-spar} combine the dense retriever with a dense lexical model that mimics sparse retrievers.
All above work focus on improving the accuracy of bi-encoder dense retrievers,
whereas our work tackles the parameter efficiency issue.

Unlike most bi-encoder dense retrievers which measure the similarity between a question and a passage 
using their corresponding \clstoken vectors,
ColBERT \cite{omar-etal-2020-colbert} develops a late-interaction paradigm and measures the similarity 
via a MaxSim operator that computes the maximum similarity between a token in a sequence and all 
tokens in the other sequence.
Such architecture has shown promising results in ODQA \cite{khattab-etal-2021-relevance}
and the BEIR benchmark \cite{santhanam-etal-2022-colbertv2}. 
Our work instead focus on the improvement on the underlying text encoders, and the MaxSim operator introduced 
by ColBERT can be applied on top of \ourmodel.

\citet{xiong-etal-2021-approximate} use the BERT-Siamese architecture for dense retrieval,
where all Transformer blocks are shared.
Compared with this architecture, \ourmodel{} is a more effective and general way to increase the parameter efficiency, by interleaving specialized Transformer blocks with shared ones.
\section{Conclusion}
We propose a new parameterization framework, \ourmodel, for improving the efficiency and robustness of dense retrieval for ODQA.
It interleaves shared encoder blocks with specialized ones in a single encoder
where some sub-networks are task-specific.
As the specialized sub-networks are sparsely activated, \ourmodel\ can provide better parameter efficiency with almost no additional computation cost.
Experiments show that 
\ourmodel\ substantially outperforms existing fully supervised bi-encoder dense retrievers on both in-domain and out-of-domain generalization.

\section{Limitations}

In this section, we point out several limitations in this work.

First, our in-domain evaluation experiments focus on passage retrieval for ODQA.
While the dense retriever is mostly successful in ODQA, it can be also used in other types of retrieval tasks which may have 
different input and output format.
For example, the KILT benchmark \cite{petroni-etal-2021-kilt} provides several knowledge-intensive tasks other than ODQA.
The performance of \ourmodel\ models trained on such retrieval tasks remain unknown.

Second, compared with traditional sparse vector models like TF-IDF and BM25, 
the cost of training is an inherent issue of dense retrievers.
Although \ourmodel\ significantly reduce the number of model parameters, the training cost is still high.

Third, in our experiments, we show that the learned routing does not outperform the deterministic routing.
This may suggest a better architecture and/or training algorithms for learned routing is needed to fully unleash the power of MoE.

Last, as observed in \S\ref{ssec:ood_eval}, there is still a gap between \ourmodel\ and BM25 in out-of-domain evaluation. 
Therefore, how to close this gap will remain a critical topic for future work on dense retrievers.

\bibliography{ref,qa}
\bibliographystyle{acl_natbib}
\clearpage

\appendix
\setcounter{table}{0}
\renewcommand{\thetable}{A\arabic{table}}

\setcounter{figure}{0}
\renewcommand{\thefigure}{A\arabic{figure}}

\section{More Routing Mechanisms}
\label{appendix:routing_mechanisms}

In the paper, only input-dependent routing is considered.
Here, we provide a more comprehensive study of routing mechanisms.
In particular, we introduce three routing mechanisms: 
the deterministic routing (\texttt{Det-R}) which is used in our main experiments,
the sequence-based routing (\texttt{Seq-R}),
and the token-based routing (\texttt{Tok-R}).
Both \texttt{Seq-R} and \texttt{Tok-R} are learned jointly with the task-specific objective.

Specifically, \texttt{Det-R} is the input-dependent routing studied in the main paper where two expert FFN sub-layers are needed for ODQA retrieval, 
one for questions and one for passages.
In this case, the router determines the expert FFN sub-layer based on whether the input is a question or a passage.

For \texttt{Seq-R} and \texttt{Tok-R}, the router uses a parameterized routing function
\begin{align}
    \texttt{R}(\uvec) = \texttt{GumbelSoftmax}(\mathbf{A} \uvec + \cvec),
\label{eqn:route_func}
\end{align}
where \texttt{GumbelSoftmax} \cite{jang-2016-gumbelsoftmax} outputs a $I$-dimensional one-hot vector 
based on the linear projection parameterized by $\mathbf{A} \in \RR^{d\times I}$ and $\cvec \in \RR^{I}$,
$I$ is the number of expert FFN sub-layers in the specialized Transformer block,
and $\uvec \in \RR^d$ is the input of the routing function.
Here, the routing function is jointly learned with all other parameters 
using the discrete reparameterization trick.
For \texttt{Seq-R}, routing is performed at the sequence level, and
all tokens in a sequence share the same $\uvec$, which is the FFN 
input vector $\hvec_{\clstoken}$ representing the special prefix token \clstoken.
For \texttt{Tok-R}, the router independently routes each token,
\ie for the $j$-th token in the sequence, $\uvec$ is set to the corresponding
FFN input vector $\hvec_j$.

For \texttt{Seq-R} and \texttt{Tok-R}, to avoid routing all inputs to the same expert FFN sub-layer, 
we further apply the entropic regularization 
\begin{align}
    L_{ent} = - \sum_{i=1}^{I} P(i)\log P(i).
\end{align}
where $P(i)=\texttt{Softmax}(\mathbf{A} \hvec + \cvec)_i$ is the probability of 
the $i$-th expert FFN sub-layer being selected. 
Hence, the joint training objective is 
\begin{equation}
    L_{joint} = L_{sim} + \beta L_{ent},
\end{equation}
where $\beta$ is a scalar hyperparameter. In our work, we fix $\beta=0.01$.

Also, all specialized Transformer blocks use the same number of expert FFN sub-layers for simplicity.

\section{Hard Negative Mining}
\label{sec:hard_negatives_mining}

Recall that in \autoref{eqn:obj_sim} the objective $L_{sim}$ needs to use a set of negative passages $\mathcal{P}$ for each question.
There are several ways to construct $\mathcal{P}$.
In \cite{karpukhin-etal-2020-dense}, the best setting uses two negative passages per question:
one is the top passage retrieved by BM25 which does not contain the answer but match most question tokens,
and the other is chosen from the gold positive passages for other questions in the same mini-batch.
Recent work shows that mining harder negative examples with iterative training can lead to better performance \cite{xiong-etal-2020-iter-train,qu-etal-2020-rocketqa}. 
Hence, in this paper, we also train \ourmodel\ with hard negatives mining.
Specifically, we first train a \ourmodel\ model with negative passages $\mathcal{P}_1$ same as \citet{karpukhin-etal-2020-dense}.
Then, we use this model to construct $\mathcal{P}_2$ by retrieving top-100 ranked passages for each question excluding the gold passage.
In the single-set training, we combine $\mathcal{P}_1$ and $\mathcal{P}_2$ to train the final model.
In the multi-set training, only use $\mathcal{P}_2$ is used to train the final model for efficiency consideration.

\begin{table}[t!]
\centering

\begin{tabular}{l|cc|cc}

\toprule
\textbf{Model} & $I$ & \textbf{\# Params} & \textbf{Dev} & \textbf{Test} \\

\midrule

DPR                                         & - & 218M     & - & 78.4  \\
\ourmodel\textsubscript{\texttt{Shared}}    & 1 & 109M   & 78.2 & 79.3\\
\ourmodel\textsubscript{\texttt{Det-R}}     & 2 & 128M & \textbf{79.2} & \textbf{80.7}  \\
\ourmodel\textsubscript{\texttt{Seq-R}}     & 2 & 128M & \textbf{79.2} & 80.6  \\
\ourmodel\textsubscript{\texttt{Seq-R}}     & 4 & 166M & 78.4 & 80.1  \\
\ourmodel\textsubscript{\texttt{ToK-R}}     & 2 & 128M & 78.5 & 79.8  \\
\ourmodel\textsubscript{\texttt{ToK-R}}     & 4 & 166M & 78.5 & 79.8  \\

\midrule
DPR$^\dagger$                                       & - & 218M     & - & 81.3  \\
\ourmodel\textsubscript{\texttt{Det-R}}$^\dagger$   & 2 & 128M & \textbf{82.4} & \textbf{83.7}  \\

\bottomrule
\end{tabular}

\caption{R@20 on NQ dev and test sets under the single-set training setting.
$I$ is the number of expert FFNs.
The \# params column shows the number of parameters in the model.
$\dagger$ means the model is trained with hard negatives mining described in \S\ref{sec:hard_negatives_mining}.
The results for DPR and DPR$^\dagger$ are reported in 
\cite{karpukhin-etal-2020-dense} and \url{https://tinyurl.com/yckar3f6}, respectively.}
\label{tab:single_set}

\end{table}

\begin{table*}[t!]
\centering

\begin{tabular}{l|cc|cc|cc|cc|cc}
\toprule

& \multicolumn{2}{c|}{\textbf{NQ}}
& \multicolumn{2}{c|}{\textbf{TriviaQA}}
& \multicolumn{2}{c|}{\textbf{WebQ}}
& \multicolumn{2}{c|}{\textbf{TREC}} 
& \multicolumn{2}{c}{\textbf{SQuAD}}
\\

\textbf{Model} 
& @20 & @100 
& @20 & @100 
& @20 & @100 
& @20 & @100 
& @20 & @100 \\
\midrule

BM25$^{(1)}$        
& 62.9 & 78.3 
& 76.4 & 83.2
& 62.4 & 75.5
& 80.7 & 89.9
& 71.1 & 81.8 \\
\midrule

\multicolumn{11}{c}{\textbf{Single-Set Training}} \\
\midrule

DPR$^{(2)}$  
& 78.4 & 85.4
& 79.4 & 85.0
& 73.2 & 81.4
& 79.8 & 89.1
& 63.2 & 77.2 \\

DPR-PAQ$^{(3)}$     
& 84.7 & {\bf 89.2}
& - & -
& - & - 
& - & -
& - & - \\

coCondenser$^{(4)}$
& 84.3 & 89.0
& 83.2 & 87.3
& - & - 
& - & -
& - & - \\

\midrule

\multicolumn{11}{c}{\textbf{Multi-Set Training (without SQuAD)}} \\
\midrule

DPR$^{(1)}$   
& 79.5 & 86.1 
& 78.9 & 84.8 
& 75.0 & 83.0
& 88.8 & 93.4
& 52.0 & 67.7 \\

DPR$^{(1)}_{\rm BM25}$  
& 82.6 & 88.6
& 82.6 & 86.5
& 77.3 & 84.7
& 90.1 & 95.0
& 75.1 & 84.4 \\

xMoCo$^{(5)}$ 
& 82.5 & 86.3 
& 80.1 & 85.7
& 78.2 & 84.8 
& 89.4 & 94.1
& 55.9 & 70.1 \\

SPAR-Wiki$^{(6)}$
& 83.0 & 88.8
& 82.6 & 86.7
& 76.0 & 84.4 
& 89.9 & 95.2
& 73.0 & 83.6 \\

SPAR-PAQ$^{(6)}$    
& 82.7 & 88.6
& 82.5 & 86.9
& 76.3 & 85.2
& 90.3 & 95.4
& 72.9 & 83.7 \\

\midrule

\multicolumn{11}{c}{\textbf{Multi-Set Training (with SQuAD)}} \\
\midrule

DPR$^\dag$
& 80.9 & 86.8
& 79.6 & 85.0
& 74.0 & 83.4
& 88.0 & 94.1
& 63.1 & 77.2 \\

DPR$^\diamond$
& 82.5 & 88.0
& 81.8 & 86.4
& 77.8 & 84.7
& 91.2 & 95.5 
& 67.1 & 79.8 \\

DPR$^\star$
& 83.7 & 88.7
& 82.6 & 86.7
& 78.9 & 85.3
& 91.6 & 95.1
& 68.0 & 80.2 \\

\ourmodel$^\diamond$
& 83.6 & 88.6 
& 82.0 & 86.6 
& 77.9 & 85.4
& 91.1 & 95.7 
& 69.7 & 81.2 \\ 

\ourmodel$^\diamond_{\rm BM25}$
& 83.8 & 88.6
& 83.3 & 87.1
& 78.7 & 85.7
& 91.6 & 95.8
& 77.2 & 86.0 \\ 

\ourmodel$^\star$
& 84.9 & {\bf 89.2} 
& 83.4 & 87.1 
& 78.9 & 85.4
& 90.8 & {\bf 96.0}
& 72.9 & 83.4 \\

\ourmodel$^\star_{\rm BM25}$
& {\bf 85.0} & {\bf 89.2}
& {\bf 84.0} & {\bf 87.5}
& {\bf 79.6} & {\bf 85.8}
& {\bf 92.1} & {\bf 96.0}
& {\bf 78.0} & {\bf 87.0} \\
\bottomrule

\end{tabular}

\caption{In-domain evaluation results.
Test set R@20 and R100 are reported.
$^{(1)}$: \cite{ma-etal-2021-dpr-replication}.
$^{(2)}$: \cite{karpukhin-etal-2020-dense}.
$^{(3)}$: \cite{oguz-etal-2021-domain}
$^{(4)}$: \cite{gao-callan-2022-unsupervised}.
$^{(5)}$: \cite{yang-etal-2021-xmoco}.
$^{(6)}$: \cite{chen-etal-2022-spar}.
$_{\rm BM25}$: combined with BM25 scores.
$^\dag$: initialized from BERT-base and without hard negatives mining.
$^\diamond$: initialized from BERT-base. 
$^\star$: initialized from coCondenser-Wiki.
The last five models are trained with the same hard negatives mining.
}
\label{tab:in_domain_eval}

\end{table*}

\begin{table}[t!]
\centering

\begin{tabular}{lc}
\toprule

\textbf{Model}   & \textbf{Num.\ Parameters} \\
\midrule

DPR      & 218M \\

coCodenser & 218M \\

xMoCo    & 218M \\

SPAR-Wiki; SPAR-PAQ       & 436M \\

DPR-PAQ  &  710M \\ 

\midrule
\ourmodel$^\diamond$; \ourmodel$^\star$ & 128M \\

\bottomrule

\end{tabular}

\caption{Number of parameters for models reported in \autoref{tab:in_domain_eval}.}
\label{tab:model_size}

\end{table}
\begin{table*}[t!]
\centering

\begin{tabular}{l|ccc}
\toprule
& Macro R@20 & Micro R@20 & Micro R@100 \\
\midrule
BM25                        & 71.2 & 70.8 & 79.2\\
\midrule
DPR\textsubscript{Multi}    & 56.7 & 56.6 & 70.1\\
\ourmodel$^\diamond$        & 64.7 & 64.3 & 76.2\\
\ourmodel$^\star$           & 66.7 & 66.2 & 77.9\\ 

\bottomrule
\end{tabular}

\caption{Out-of-domain evaluation results on EntityQuestions. 
We report macro R@20 scores which are used in \cite{sciavolino-etal-2021-entityq} 
as well as micro R@20 and R@100 scores which are used in \cite{chen-etal-2022-spar}.
Results for BM25 and DPR\textsubscript{Multi} are from \cite{sciavolino-etal-2021-entityq} and \cite{chen-etal-2022-spar}.}
\label{tab:ood_entityq}

\end{table*}

\section{Comparing \ourmodel\ Variants}
\label{ssec:single_set}

In this part, we compare different \ourmodel\ variants discussed in \S\ref{appendix:routing_mechanisms} 
by evaluating their performance on NQ under the single-set training setting.
We use the bi-encoder dense passage retriever (DPR) from \cite{karpukhin-etal-2020-dense} as our baseline.
All models including DPR are initialized from the BERT-base \cite{devlin-etal-2019-bert}.\footnote{Without further specification, we only consider the uncased version throughout the paper.}
All \ourmodel\ models are fine-tuned up to $40$ epochs with Adam \cite{kingma2014adam} using 
a learning rate chosen from \{$3e-5, 5e-5$\}.
Model selection is performed on the development set following \cite{karpukhin-etal-2020-dense}.
Results are summarized in \autoref{tab:single_set}.

\ourmodel\textsubscript{Shared} is a variant without any task-aware specialization, \ie there is a single expert FFN sub-layer in the specialized Transformer block and the router is a no-op.
As shown in \autoref{tab:single_set}, it outperforms DPR while using only 50\% parameters.

Task-aware specialization brings extra improvements, with little increase in model size.
Comparing the two learned routing mechanisms, \texttt{Seq-R} achieves slightly better results than \texttt{Tok-R},
indicating specializing FFNs based on sequence-level features such as sequence types is more effective for ODQA dense retrieval.
This is consistent with the positive results for \texttt{Det-R}, which consists of two expert FFNs specialized for
questions and passages, respectively.
We also find that adding more expert FFNs does not necessarily bring extra gains, 
and $I=2$ is sufficient for NQ.
Consistent with the results on DPR, the hard negatives mining described in \S\ref{sec:hard_negatives_mining} can 
further boost \ourmodel\textsubscript{\texttt{Det-R}} performance by 3.0 points in test set R@20.
Since \texttt{Det-R} achieves the best R@20,
our subsequent experiments focus on this simple and effective specialization strategy.
In the remainder of the paper, we drop the subscript and simply use \ourmodel\ to denote models using \texttt{Det-R}.

\section{Details about In-Domain Evaluations}
\label{appendix:in_domain_evaluatios_details}

All \ourmodel\ models are fine-tuned up to $40$ epochs with Adam \cite{kingma2014adam} using 
a learning rate chosen from \{$3e-5, 5e-5$\}.
In our experiments, hard negatives are mined from NQ, TriviaQA and WebQ.
We combine NQ and TriviaQA development sets for model selection.

We also present results of hybrid models that linearly combine the dense 
retrieval scores with the BM25 scores, 
\begin{equation}
\text{sim}(\qvec, \pvec) + \alpha \cdot \text{BM25}(\qvec, \pvec).
\end{equation}
We search the weight $\alpha$ in the range $[0.5, 2.0]$ with an interval of $0.1$
based on the combined development set mentioned above.
Unlike \cite{ma-etal-2021-dpr-replication},
we use a single $\alpha$ for all five datasets instead of dataset-specified weights
so that the resulting hybrid retriever still complies with the multi-set setting in a strict sense.
The same normalization techniques described in \cite{ma-etal-2021-dpr-replication} is used.
Similar to \cite{karpukhin-etal-2020-dense,ma-etal-2021-dpr-replication},
we separately retrieve $K^\prime$ candidates from \ourmodel\ and BM25,
and then retain the top $K$ based on the hybrid scores,
though we use a smaller $K^\prime = 100$.

We used 16 V100-32GB GPUs and it took 9 hours to train our models.

\section{Dataset Licenses and Intended Use}

All datasets used in our experiments are English datasets.
The datasets used in this paper are released under the following licenses.

\begin{itemize}
\item NaturalQuestions \cite{kwiatkowski-etal-2019-natural}: CC-BY-SA 3.0 License
\item TriviaQA \cite{joshi-etal-2017-triviaqa}: non-commercial research purposes only
\item WebQuestions \cite{berant-etal-2013-semantic}: CC-BY 4.0 License.
\item SQuAD \cite{squad1}: CC-BY-SA 4.0 License
\item EntityQuestions \cite{sciavolino-etal-2021-entityq}: MIT License
\item ArguAna \cite{wachsmuth-etal-2018-retrieval}: not specified
\item DBPedia \cite{hasibi-etal-2017-dbpedia}: not specified
\item FEVER \cite{thorne-etal-2018-fever}: license terms described on the applicable Wikipedia article pages, and CC BY-SA 3.0 License

\item HotpotQA \cite{yang-etal-2018-hotpotqa}: CC BY-SA 4.0
\end{itemize}

Our use is consistent with their intended use.

\end{document}